\documentclass[letterpaper, 10 pt, conference]{ieeeconf}

\usepackage{graphics}
\usepackage{epsfig}
\usepackage{mathptmx}
\usepackage{times}
\usepackage{amsmath}
\usepackage{amssymb}
\usepackage{color}
\usepackage{gensymb}

\title{\LARGE \bf Deep Reinforcement Learning for Motion Planning of Mobile Robots}

\author{ Leonid Butyrev{\small$^{1}$}, Thorsten Edelh\"au{\ss}er{\small{$^{1}$}} and Christopher Mutschler{\small{$^{1,2}$}}\\
		{\tt\small \{ butyreld | thorsten.edelhaeusser | christopher.mutschler \} @iis.fraunhofer.de}\vspace{3pt}\\
		$^1$Fraunhofer Institute for Integrated Circuits IIS, Precise Positioning and Analytics Department,\\
		Machine Learning and Information Fusion Group, Nuremberg, Germany\vspace{2pt}\\
		$^2$Friedrich-Alexander-University Erlangen-Nuremberg (FAU), Computer Science Department\\
		Machine Learning and Data Analytics Lab, Erlangen, Germany
}

\begin{document}

\maketitle
\thispagestyle{empty}
\pagestyle{empty}

\begin{abstract}
This paper presents a novel motion and trajectory planning algorithm for nonholonomic mobile robots
that uses recent advances in deep reinforcement learning. Starting from a random initial state, i.e.,
position, velocity and orientation, the robot reaches an arbitrary target state while
taking both kinematic and dynamic constraints into account. Our deep reinforcement learning agent
not only processes a continuous state space it also executes continuous actions, i.e., the
acceleration of wheels and the adaptation of the steering angle.

We evaluate our motion and trajectory planning on a mobile robot with a differential drive in a
simulation environment.
\end{abstract}

\section{Introduction}
\label{section:introduction}

Motion planning is a well-known fundamental challenge in mobile robotics~\cite{choset2005principles}.
Depending on the dynamic nature of the environment and the physical constraints that are considered
it soon becomes complex. Moreover, in many real world applications mobile robots are faced with
highly dynamic environments and frequent changes in immediate surroundings that require quick
reaction times.

\begin{figure}[t!]%
	\centering%
	\includegraphics[width=0.6\linewidth]{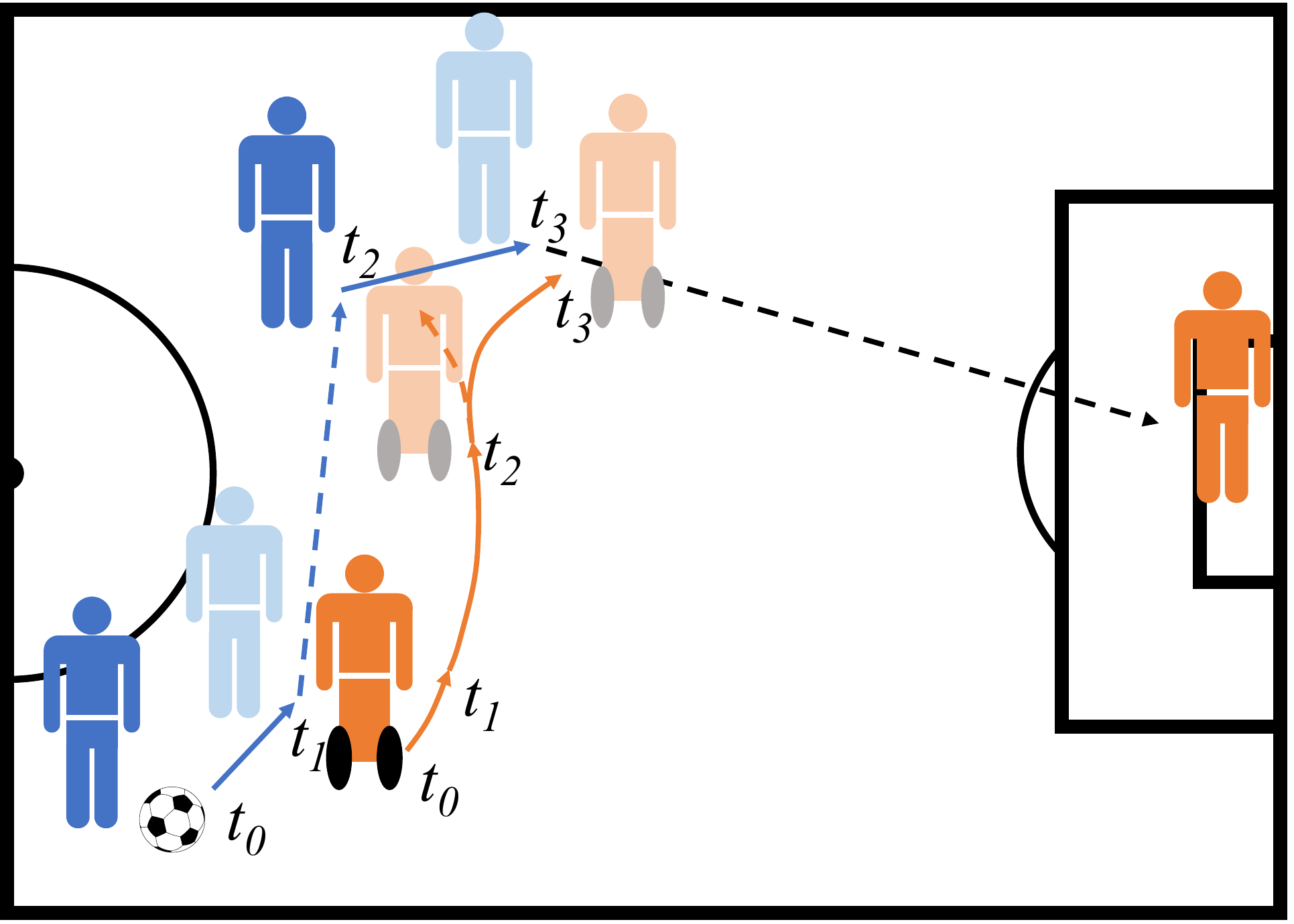}%
	\vspace{-2mm}%
	\caption{Mobile robots interact with humans in a dynamic soccer application.}%
	\label{fig:application}%
	\vspace{-6mm}%
\end{figure}

Consider the soccer application depicted in Fig.~\ref{fig:application}. At least one player (blue)
tries to score a goal while at least one mobile robot (orange) defends. A real-time locating system
(RTLS) tracks the ball and the player kinematics and the mobile robots and delivers a stream of
their positions~\cite{Feigl2018,Niitsoo2019}. A high-level tactical analysis monitors all trajectories and generates a stream of
target positions $t_0, t_1...$ per robot so that the goal is best shielded at any time $t$. As
inertia limits the robots' movement flexibility we need to keep them constantly be moving.
Therefore, a tactical analysis monitors the players' movement patterns and also generates
appropriate orientations and velocities for the prospective target positions. These together describe
a target state that is then used to estimate a trajectory for the robot (even before it reaches
its current target state, see $t_2$).

Sampling methods~\cite{lavalle2001randomized} estimate such trajectories by
reducing the continuous control problem to a discrete search problem. But the reduction in
complexity comes with a loss in optimality and completeness of the solution which in turn can lead
to unsatisfying results in such highly dynamic environments~\cite{lavalle2001randomized}. However,
while it is possible to find such a trajectory through analytical methods~\cite{lim2018hierarchical}
it becomes significantly more challenging if we also consider the kinematic and dynamic side
constraints. This leads to a non-linear optimization problem that can no longer be solved
analytically fast enough for such interactive scenarios.

To address the full motion estimation problem we use recent advances in deep reinforcement learning
and teach a robot to generate feasible trajectories that obey kinematic and dynamic constraints in
real-time. A simulation environment models the environmental and physical
constraints of a real robotic platform. We generate random target states and let
a deep reinforcement learning (RL) agent based on Deep Deterministic Policy Gradient
(DDPG)~\cite{lillicrap2015continuous} interact with the simulator on a trial-and-error basis to solve
the motion planning problem within a continuous state and action space.

The remainder of this paper is structured as follows. Sect.~\ref{section:related_work} discusses
related work and Sect.~\ref{section:background} provides background information. Next, we show our
main contributions: we formalize the problem and describe the implementation of our reward-system
and RL agent in Sect.~\ref{section:method}. Sect.~\ref{section:evaluation} evaluates our approach
with a deep RL-agent
within a simulation environment that also models the robot's and
environmental physics before Sect.~\ref{section:conclusion} concludes.

\section{Related Work}
\label{section:related_work}

Trajectory planning~\cite{gasparetto2012trajectory} generates a sequence of control inputs based on
a geometric path and a set of kinematic or dynamic constraints. The geometric path comes from path
planning methods. Both together solve the motion planning problem. While they are typically
decoupled~\cite{choset2005principles}
some approaches solve the motion planning problem directly~\cite{choset2005principles}.

\textbf{Sampling-based approaches} discretize the action and state space, and transform the
underlying control problem into a graph search problem~\cite{likhachev2008anytime},
which allows for the use of precomputed motion
primitives~\cite{howard2008state,pivtoraiko2009differentially,pivtoraiko2011kinodynamic}. Further
extensions propose better-suited cost functions~\cite{lavalle2006planning} for the search
algorithms, e.g. state lattice planners~\cite{wang2015state,ziegler2009spatiotemporal} sample the
state space with a deterministic lattice pattern. 
Randomness~\cite{lavalle2001randomized}
helps to reduce generated trajectories that are not relevant for the given search problem.
While sampling-based approaches are not complete, they offer probabilistic
completeness~\cite{hsu2006probabilistic,lavalle2001randomized}, i.e., the
probability to find a solution grows with the runtime. 
However, as sampling based approaches discretize the state and action space, they require a high amount
of computational resources~\cite{lim2018hierarchical} to account for
all possible motions. In general, a higher number of motion primitives leads to a
rapid growth in time complexity~\cite{lindemann2005current}. Realistic motion planners need
thousands of such primitives and are therefore not suitable for a
interaction~\cite{lindemann2005current}.

\textbf{Interpolation-based methods} use way points to compute a path with higher trajectory
continuity. In the simplest case they interpolate with lines and
circles. More elaborate approaches use clothoid curves~\cite{brezak2014real},
polynomial curves~\cite{glaser2010maneuver} and B\'ezier curves~\cite{rastelli2014dynamic}. Clothoid
curves allow for the definition of trajectories based on linear changes in curvature since their
curvature is defined as their arc-length~\cite{brezak2014real}. Polynomial curves are commonly used
to also take side-constraints into account, since their coefficients are defined by the constraints
in their beginning and ending segments~\cite{glaser2010maneuver}. B\'ezier curves are defined by a
set of control points and represent parametric curves~\cite{rastelli2014dynamic}. However, the
trajectories are not necessarily optimal and although these methods may generate smooth
trajectories they may also not obey kinodynamic side-constraints.

\textbf{Numerical optimization} extends both sampling- and interpolation-based algorithms. Kinematic
and dynamic constraints require additional optimization. Numerical
optimization generates trajectories based on a differentiable cost function and side-constraints.
Under convex cost and constraint functions~\cite{lim2018hierarchical} they find globally optimal
trajectories~\cite{ziegler2014trajectory}. Either they optimize a suboptimal
trajectory~\cite{dolgov2010path} or compute a trajectory on predefined
constraints~\cite{ziegler2014making}. Such planners generate continuous trajectories by optimizing a
function that considers planning parameters like position, velocity, acceleration and
jerk~\cite{ziegler2014making,ziegler2014trajectory}. However, the downside of the additional
optimization step is the rise in time complexity. While a suboptimal trajectory can be computed
quickly, an optimal solution is time consuming~\cite{kalakrishnan2011stomp} and therefore less
applicable for time-critical tasks.

Since we need a fast and frequent (re-)computation of (ideally near-optimal) trajectories that obey
kinodynamic side-constraints existing methods are not sufficient. Instead, our deep reinforcement
learning is trained offline and delivers trajectories satisfying all side-constraints in real-time
given a continuous action and state space.

\section{Background on Reinforcement Learning}
\label{section:background}

\begin{figure}[t!]%
	\centering%
	\includegraphics[width=0.7\linewidth]{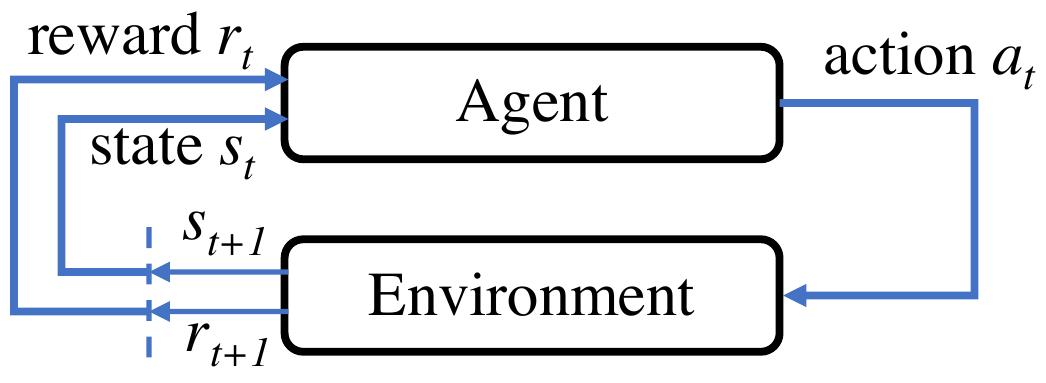}%
	\vspace{-4mm}%
	\caption{The basic reinforcement learning scenario.}%
	\vspace{-6mm}%
	\label{fig:RLstandard}%
\end{figure}

The key idea behind reinforcement learning is an agent that interacts with the environment, see
Fig.~\ref{fig:RLstandard}. In each time step the agent receives an observation $s_t$, i.e., some (partial)
information on the current state, and selects an action $a_t$ based on the observation. The
environment then rewards or punishes this action with a reward $r_{t+1}$.

\subsection{Markov Decision Process}

A reinforcement learning task that satisfies the Markov property, i.e, that the probabilities of
future states $s_{T>t}$ only depend on the state $s_t$ but not on previous events, is
called a Markov Decision Process (MDP). With a set of states $s \in S$, actions
$a \in A$, and rewards $r \in R$ a controlled process with Markov dynamics at time $t=0,1,...$ is
defined by
\begin{align*}
p(s^{\prime}  | s, a) &= \text{Pr}(s_{t+1}=s^\prime | s_t=s,a_t=a) \text{, and}\\
r(s,a,s^\prime) &= \mathbb{E} [ R_{t+1} | s_{t}=s, a_{t}=a, s_{t+1}=s^\prime ],
\end{align*}
where $p(s^\prime)$ is defined by a probability distribution that models the transition dynamics (as
often state transition may be probabilistic) and $r(s,a,s^\prime)$ is defined by the expected reward if
we choose action $a_t$ in $s_t$ and end up in $s_{t+1}$.

In fully observed environments an observation $o_t$ completely
describes the underlying (real) state of the environment $s_t$. However, in partially observable
environments the agent must estimate the real state $s_t$ based on $o_t$ or a set of past and
present observations. For simplification we consider a fully observable environment, i.e., we
assume $o_t = s_t$.

\subsection{Reinforcement Learning}

We use the state-value function $v^\pi(s)$ to denote the value of state $s$ under a given policy $\pi$,
i.e., the expected total reward given that the agent starts from state $s$ and behaves according to
$\pi$. Accordingly, the value of a state is defined as the (expected) sum of discounted future
rewards
\begin{center}
\begin{math}
\displaystyle
v_{\pi}(s) = \mathbb{E}_{\pi} \left[ G_t | S_t = s \right] =
\mathbb{E}_{\pi} \left[ \sum_{k=0}^{\infty} \gamma^{k}R_{t+k+1} | S_t = s \right],
\end{math}
\end{center}
where $\mathbb{E}\left[\cdot\right]$ denotes the expected value, $t$ is any time step, and
$\gamma \in \left[0,1\right]$ is the discount factor that favors immediate rewards over future
rewards (and that also determines how far in the future the rewards are considered). The goal of
reinforcement learning lies in the maximization of the expected discounted future reward from the
initial state.

The policy $\pi:$ $S \rightarrow P(A)$  maps a state $s \in S$ to a probability distribution over
the available actions $a \in A$. Since the future actions of
the agent are defined by the policy $\pi$ and the state transitions by the transition
dynamics $p(s_{t+1}|s_t,a_t)$ we can estimate the future reward that we can expect in any given
state.

Similarly, we can define the value of taking action $a$ in state $s$ under a policy $\pi$ as the 
expected return starting from $s$, taking the action $a$, and thereafter following policy $\pi$:
\begin{align*}
	q_{\pi}(s,a) &= \mathbb{E}_{\pi} \left[ G_t | S_t=s, A_t=a \right]\\
	&= \mathbb{E}_{\pi} \left[ \sum^{\infty}_{k=0} \gamma^{k}R_{t+k+1} | S_t = s, A_T=a \right],
\end{align*}
which is called the (state-)action-value function for policy $\pi$. The recursive formula describes the 
relationship between the value of a state and the value of the successor states and is also often
called \textit{Bellman equation for $v_\pi$}.

To learn the optimal action-value function is one of the central goals of reinforcement learning.
Popular methods are dynamic programming, on- and off-policy monte-carlo-methods,
temporal-difference learning, and Q-learning.

\subsection{Deep Deterministic Policy Gradient (DDPG)}

Value-based reinforcement learning such as Q-learning has poor convergence properties as slow
changes in the value estimate have a big influence on the policy. Instead, policy-based
reinforcement learning directly updates the policy but often converges to local optima.

\begin{figure}[b!]%
	\vspace{-5mm}%
	\centering%
	\includegraphics[width=0.8\linewidth]{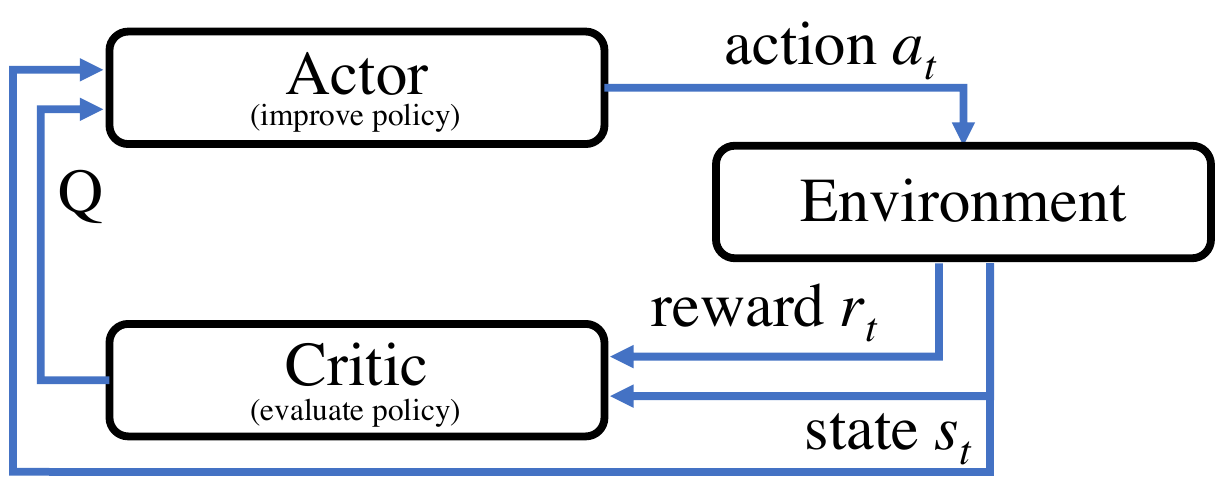}%
	\vspace{-3mm}%
	\caption{Actor-critic framework.}%
	\label{fig:actor-critic}%
\end{figure}

Actor-critic-methods combine the best of both worlds, see Fig.~\ref{fig:actor-critic}. Instead of
directly estimating the state-action-value function we use two separate components. While a
critic estimates the value of state-action-pairs an actor takes a state $s_t$ and estimates the
best action from within this state.

The combination of an actor-critic-framework and deep neural networks allows the application of
Q-learning 
to a continuous state and action space~\cite{lillicrap2015continuous}.
DDPG uses a parameterized actor-function $\mu(s|\theta^{\mu})$ and critic-function $Q(s,a|\theta^Q)$,
where $\theta^{\mu}$ and $\theta^{Q}$ are the weight parameters of neural networks that
approximate these functions.

The critic function is learned using the the Bellman equation and standard back-propagation. As
learning through neural networks in RL is unstable DDPG uses a concept of target networks, i.e., 
 $Q'$ and $\mu'$, to ensure that the critic's network updates only change slowly. The target
 networks are initialized by copying the actor and critic networks and then updated using
 $\tau \ll 1$ through
 \begin{align*}
 \theta' &\leftarrow \tau\theta + (1-\tau)\theta' \text{, and}\\
 \mu' &\leftarrow \tau\mu + (1-\tau)\mu'.
 \end{align*}

To train the actor we use the chain rule to the expected return from the start distribution
with respect to the actor parameters. To better explore the action space we sample from a normal
distribution and add it to the actor policy.

To allow for minibatch (learning from a batch of transitions) and off-policy learning, as well
as the i.i.d assumption (independent and identically distributed), DDPG uses a replay
buffer that stores transitions $(s_t,a_t,r_t,s_{t+1})$ after each interaction from which we later
sample for training.

\section{Method}
\label{section:method}

\subsection{Kinematic Model}

Although our application assumes a differential drive mobile robot we model the robot by a unicycle with a
clearer physical interpretation as both kinematics are
equivalent~\cite{siciliano2009robotics}. Hence, its configuration is given by
$q = [x\ y\ \theta]^T$, with wheel orientation $\theta$ and the contact point of the wheel with
the ground $(x,y)$. The kinematic model is described by
\begin{align*}
\begin{bmatrix}
\dot{x}\\\dot{y}\\\dot{\theta}
\end{bmatrix} & = 
\begin{bmatrix}
cos \theta\\sin \theta\\0
\end{bmatrix} \nu + 
\begin{bmatrix}
0\\0\\1
\end{bmatrix} \omega,
\end{align*} 
where $\nu$ is the velocity and $\omega$ is the steering velocity.


\subsection{Dynamic Constraints}

We also consider physical limitations known from real physical robot platforms. We limit the linear
acceleration $\Delta\nu$ and the angular acceleration $\Delta\omega_{target}$ by
\begin{align*}
|\Delta\nu| &\le a_{max,linear} \cdot dt\text{, and}\\
|\Delta\omega_{target}| &\le a_{max,angluar} \cdot dt
\end{align*}
in each time step $dt$. As we need to prevent the robot from tipping over we also include centrifugal
force limits:
\begin{align*}
|\nu_{target} \cdot \omega_{target}| \le a_{max,lateral}.
\end{align*}

We also restricted the steering to a maximum velocity $|\nu| \le \nu_{max}$ and an angular rate
$|\omega| \le \omega_{max}$.

\subsection{Problem Formulation}
\begin{figure}[t!]%
	\centering%
	\includegraphics[width=0.5\linewidth]{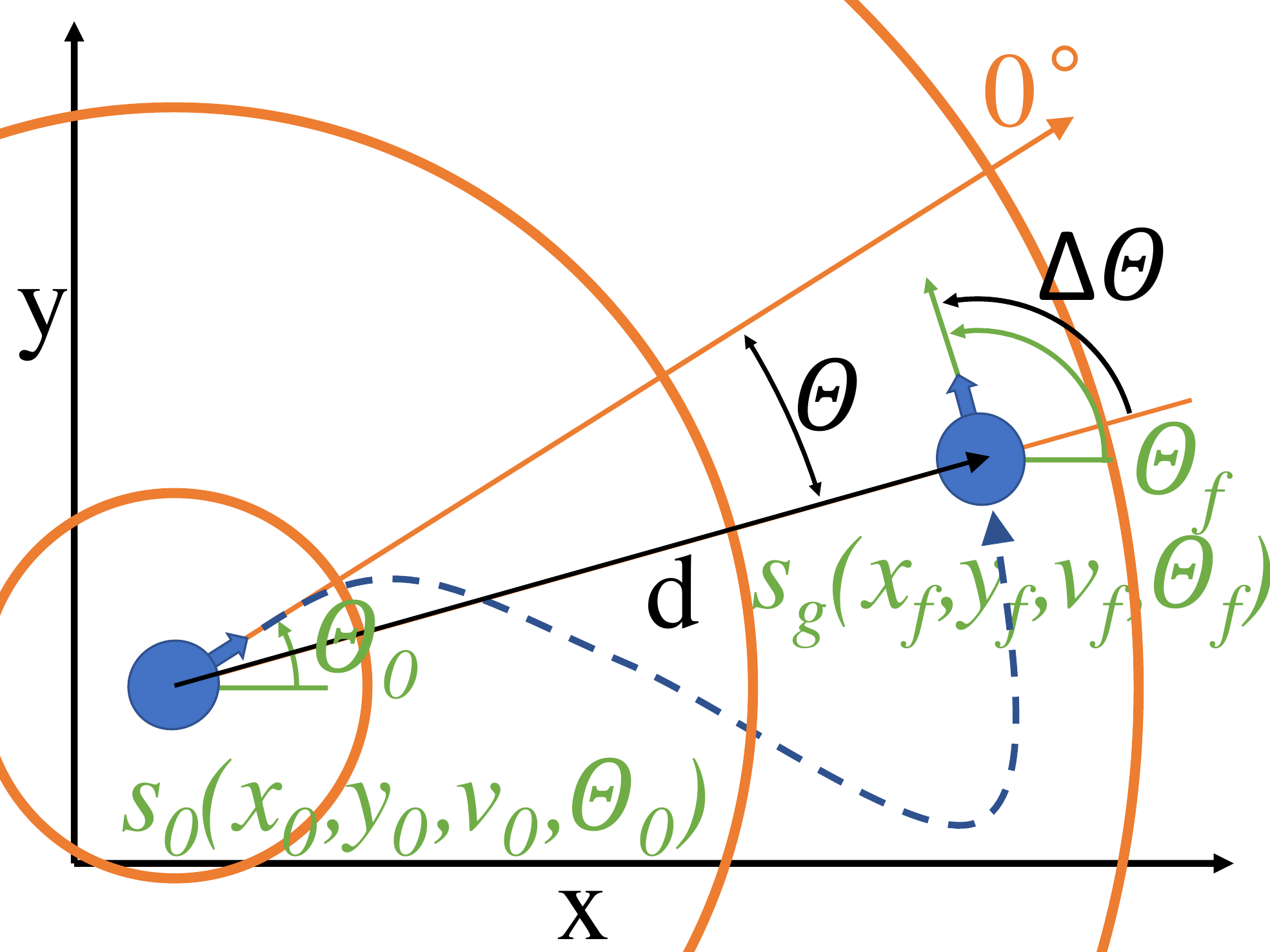}%
	\vspace{-3mm}%
	\caption{Formulization of the motion planning problem.}%
	\vspace{-6mm}%
	\label{fig:problem}%
\end{figure}%

In accordance to our kinematic model we describe a state $s$ by $(x,y,\nu,\theta)^T$, where $x$ and $y$
are the robot's Cartesian coordinates, $\nu$ its linear velocity, and $\theta$ its forward
orientation.

To formalize our motion planning consider Fig.~\ref{fig:problem}. The robot's task is to reach a
goal state $s_g$ from a start state $s_0$. We consider a control affine non-linear dynamic system
with drift dynamics and stochastic effectiveness of control inputs. Both are unknown, but are assumed
to be locally Lipschitz functions, i.e., locally constant. We also
assume the control effectiveness function to be bounded.


%

Our RL-Agent receives a new observation $o_t$ and a reward $r_t$ per time-step $t$ for the
current and past actions from our environment. We define an observation by
$o_t = \big(d, \theta, \Delta\nu, \Delta\theta, \nu, \phi \big)$,
where $(d, \theta)$ is the polar coordinate of the goal position in the robot's frame,
$\Delta\nu$ is the residual of the current to the goal velocity, $\Delta\theta$ is the
residual of the (global) current to the goal orientation, and $(\nu, \phi)$ are the actual
velocity and rotation rate of the robot. The latter two ensure a fully observable environment to
our agent.\footnote{Those can be also omitted if we add a memory to the agent, e.g. with a
recurrent neural network~\cite{Heess15} or if we stack some past observations together.}

We relax the motion planning constraints and allow the robot to reach the goal-state both
forwards and backwards as long as its motion vector is correct. Therefore, the residual of the velocity
$\Delta\nu$ is $\nu_f - |\nu|$ and
$$\Delta\theta = \theta_f - \left\{\def\arraystretch{1.2}\begin{tabular}{@{}l@{\quad}l@{}}
  $\theta$ & if $\nu \ge 0$ \\
  $\theta + \pi$ & if $\nu < 0$.
\end{tabular}\right.$$
Afterwards we normalize $\Delta\theta$ to $]$$-\pi, \pi$$]$. 

Our simulation environment returns an immediate reward $r_t = \frac{1}{1+e}$, where the error
$e = \left|\left| \hat{s}_g - \hat{s}_t\right|\right|_2$ is the Euclidean distance between the
current state $\hat{s_t}$ and the goal state $\hat{s}_g$. 

We end an episode, if either the error is lower than a threshold $e < \varepsilon$ or if we reach
a maximum number of steps $t \ge T$. In the former case, the agent receives an additional fixed
reward $R$ at the last step to enforce this behavior as the reward alone does not tell the agent
how time-efficient its trajectory has been.

\begin{figure}[t!]%
	\centering%
	\includegraphics[width=0.8\linewidth]{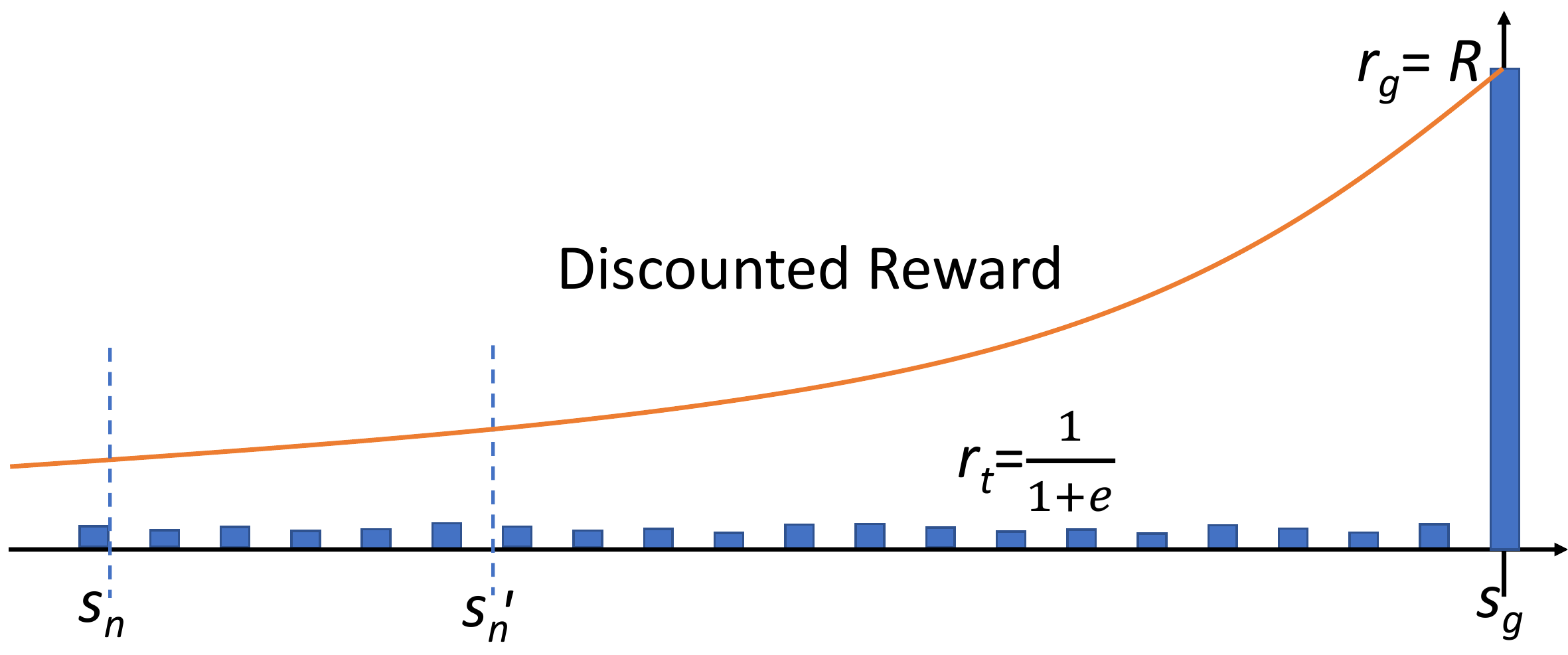}%
	\vspace{-3mm}%
	\caption{Discounted Reward.}%
	\vspace{-5mm}%
	\label{fig:discounted}%
\end{figure}

Fig.~\ref{fig:discounted} shows how a discounted reward (orange line) helps to score for
time-optimality. The closer a state is to our final goal state $s_g$ the higher is the discounted
reward for the agent (e.g. it is higher at $s_n'$ than at $s_n$). This encourages the agent to move
closer to $s_g$ at a later point. The orange curve is mainly influenced by the final reward
$r_g = R$ if the agent has reached the goal state $s_g$. If the agent does not reach the goal state
the orange curve uses the discounted immediate reward $r_t$, which is much smaller. However,
this smaller reward initially helps the agent to find the goal state in the search space.

\section{Evaluation}
\label{section:evaluation}

We use a simulator (Sect.~\ref{section:evaluation:simulation}) to train our
RL-agent (Sect.~\ref{section:evaluation:parameters}). We evaluate the optimization of our agent
during training (Sect.~\ref{section:evaluation:training}) before we show the efficiency of our
RL-based approach (Sect.~\ref{section:evaluation:accuracy}). We also evaluate our approach in a
complex use-case scenario (Sect.~\ref{section:evaluation:application}).

\subsection{Simulation and Training Setup}
\label{section:evaluation:simulation}

To train and test our RL-based motion planner we implemented the environment, i.e., the kinodynamic
constraints, the kinematic model, our reward system, and the action, observation and state scheme,
within the OpenAI Gym framework~\cite{openAIGym2016}. We implemented a discrete form with a
step-size of $dt = 0.1s$ and applied the following kinematic constraints:
$a_{max,linear}$=$2.2 \frac{m}{s^2}$, $a_{max,angular}$=$2.0 \frac{m}{s^2}$,
$a_{max,lateral}$=$1.0 \frac{m}{s^2}$, $\nu_{max}$=$4.0 \frac{m}{s}$ and
$\omega_{max}$=$4.5 \frac{1}{s}$.

To let our agent learn, we generate random pairs of start- and goal-states. We transform them such
that the start-state is at the origin $(0,0)$ and its initial orientation points to the positive
x-axis, and sample its velocity ($0...v_{max}$) from a uniform distribution. We generate the
goal-state in relation to the local frame within a position sampled uniformly within a distance of
$]0.5m ... 5.0m]$ to the origin. We also sample the goal orientation and velocity uniformly. Note,
that we allow to reach the goal state backwards or forwards, i.e., the orientation then turns by
$180^o$ with a negative velocity.

The agent gets a reward $R$=$100$ 
if it reaches the final goal-state within a threshold of $e < 0.5$. We used a grid search
to evaluate the reward $R \in \{1, 10, 100\}$. We end an episode if the agent needs more
than $200$ steps to reach the goal. In this case, the last reward is the immediate reward $r_t$.

For both training and testing the agent and the simulator run on a desktop machine equipped
with an Intel Core i7-7700 CPU@3.60GHz (4 cores, 8 threads), 32GB memory, and an Nvidia
GeForce GTX1070 with 8GB memory. We implemented all our algorithm in Python.

\subsection{Training Parameters of the RL-Agents}
\label{section:evaluation:parameters}

Our DDPG agent uses two separate networks for the actor and the critic that have a similar design. Both use
an input layer, three fully connected hidden layers (with 200 neurons each) and an output layer. The
hidden layers are necessary to approximate the non-linearities in the value-function.
Both the actor and the critic receive the observation vector $o_t$ at the input layer. The actor output
is additionally connected to the second hidden layer of the critic. We use a linear action (ReLU) for the critic's
output layer and \texttt{tanh}-activation in
all other cases. While we initialize all biases with a constant value of $0.1$, we initialize the
weights of the critic by sampling from a normal distribution
$\mathcal{N}$($\mu$=$0$, $\sigma^2$=$0.1$) and the actor by sampling from
$\mathcal{N}$($\mu$=$0$, $\sigma^2$=$0.3$).

For the training using back-propagation we use the ADAM optimizer. The
exploration in our DDPG agent uses a simple $\epsilon$-greedy algorithm. We evaluated the following
combinations with a grid search to find the best hyperparameters for the training (best setting in
bold): learning rate actor ($\alpha \in \{10^{-1}, \mathbf{10^{-2}}, 10^{-3}, 10^{-4}, 10^{-5}\}$,
$\beta_1$=$0.9$, $\beta_2$=$0.999$, $\epsilon$=$10^{-8}$) learning rate critic
($\alpha \in \{10^{-3}, \mathbf{10^{-4}}, 10^{-5}, 10^{-6}, 10^{-7}\}$, $\beta_1$=$0.9$, $\beta_2$=$0.999$,
$\epsilon$=$10^{-8}$), discount factor $\gamma \in \{0.5, 0.8, 0.9, \mathbf{0.95}\}$, batch size
$\in \{ \mathbf{500}, 1000, 2000\}$ (has no influence since it's indirect proportional to the training
time), exploration probability $\epsilon \in \{\mathbf{0.5}, 0.75\}$, exploration variation
$\sigma \in \{\mathbf{3.0}, 5.0\}$, memory size ($50000$), and soft update factor $\tau = 0.1$.

For the test we set $\epsilon$=$0$ to fully exploit the policy. While the immediate output of
the actor is a greedy action we select a non-greedy action by sampling from a normal distribution
with the greedy action as its mean: $a'$=$\mathcal{N}(a,\sigma)$. The variance in training defines
the randomness of the action: with a higher variance we more likely choose random actions.


\subsection{Optimization during Training}
\label{section:evaluation:training}

\begin{figure}[t!]
	\centering
	\includegraphics[width=0.95\linewidth]{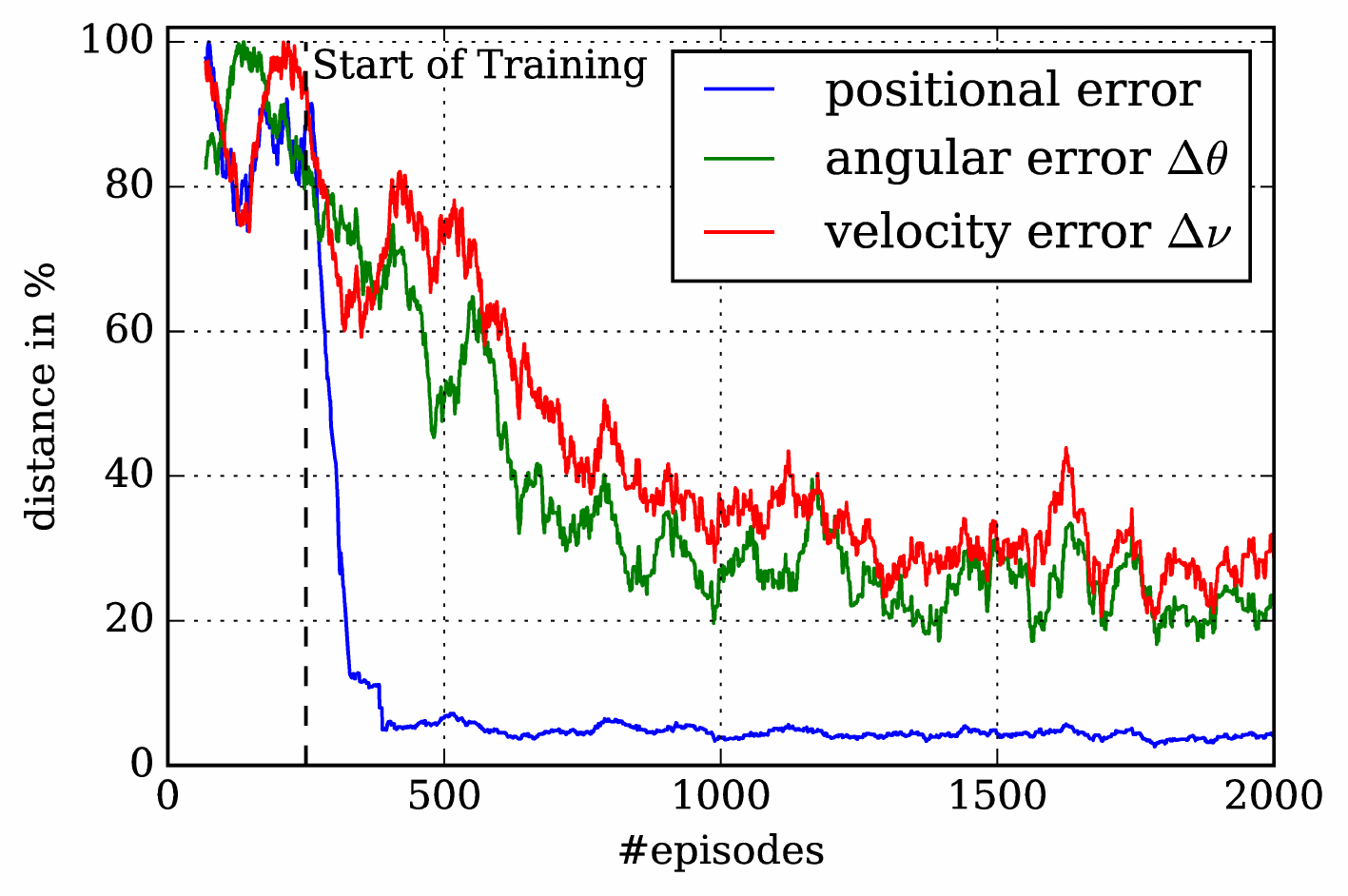}
	\vspace{-0.4cm}
	\caption{Distance per dimension on training for the 4-dimensional problem.}
	\vspace{-0.4cm}
	\label{fig:distance4D}
\end{figure}

To understand how our agent optimizes the motion planning we train the agent in the 4D state space
and analyze the error on the particular dimensions over the training episodes.

Fig.~\ref{fig:distance4D} shows the distance/error of the 4D-agent per state dimension, i.e.,
positional distance, angular distance $\Delta\theta$, and velocity distance $\Delta\nu$, over the
training episodes. As we now only focus on the optimization of each individual dimension we scale
the y-axis to show the results in percentages (the maximal positional distance was $14.0$m,
the maximal angular distance was $179.7\degree$, and the maximal velocity distance was $3.88$m/s).
The episodes that the agent run until the dashed line at episode \#250 are (until that point) only
used to fill the replay memory. They are generated using the randomly initialized policy. 

Afterwards, the agent samples from the replay memory and starts training with a absolute positional
error of around $14$m, and angular error of approx. $90\degree$ and a velocity error of $1.7$m/s
(max. $4.0$m/s allowed). The agent optimizes for the positional error first. It drops below $1.0$m
after only 500 episodes, i.e., 250 episodes in training. The optimization of the angular error comes
with a little delay: it drops below $40\degree$ at episode 1,000. The agent optimizes for the
velocity at latest. Starting with an error of $1.7$m/s the error drops below $0.6$m/s after 1,700
episodes. Interestingly, velocity and angular distance fight for a minimum error until the end. Note
that the errors are higher during training as the agent applies a non-greedy behavior.




\subsection{Accuracy Results}
\label{section:evaluation:accuracy}

The agent's task is to reach a goal-state with time- and way-efficient controls. However, as the
4D problem is complex we break it down to subproblems on which we analyze our agent's behavior.

We separately train an agent and evaluate its accuracy within the 2D- (only position), 3Da-
(position and orientation), 3Db- (position and velocity), and 4D-problem (complete goal-state).
Hence, we trained four agents using the same reward function for 4,000 episodes each. The total
training time has been 45 minutes for all agents in parallel.
\begin{figure}[b!]%
	\centering%
	\vspace{-4mm}%
	\includegraphics[width=0.8\linewidth]{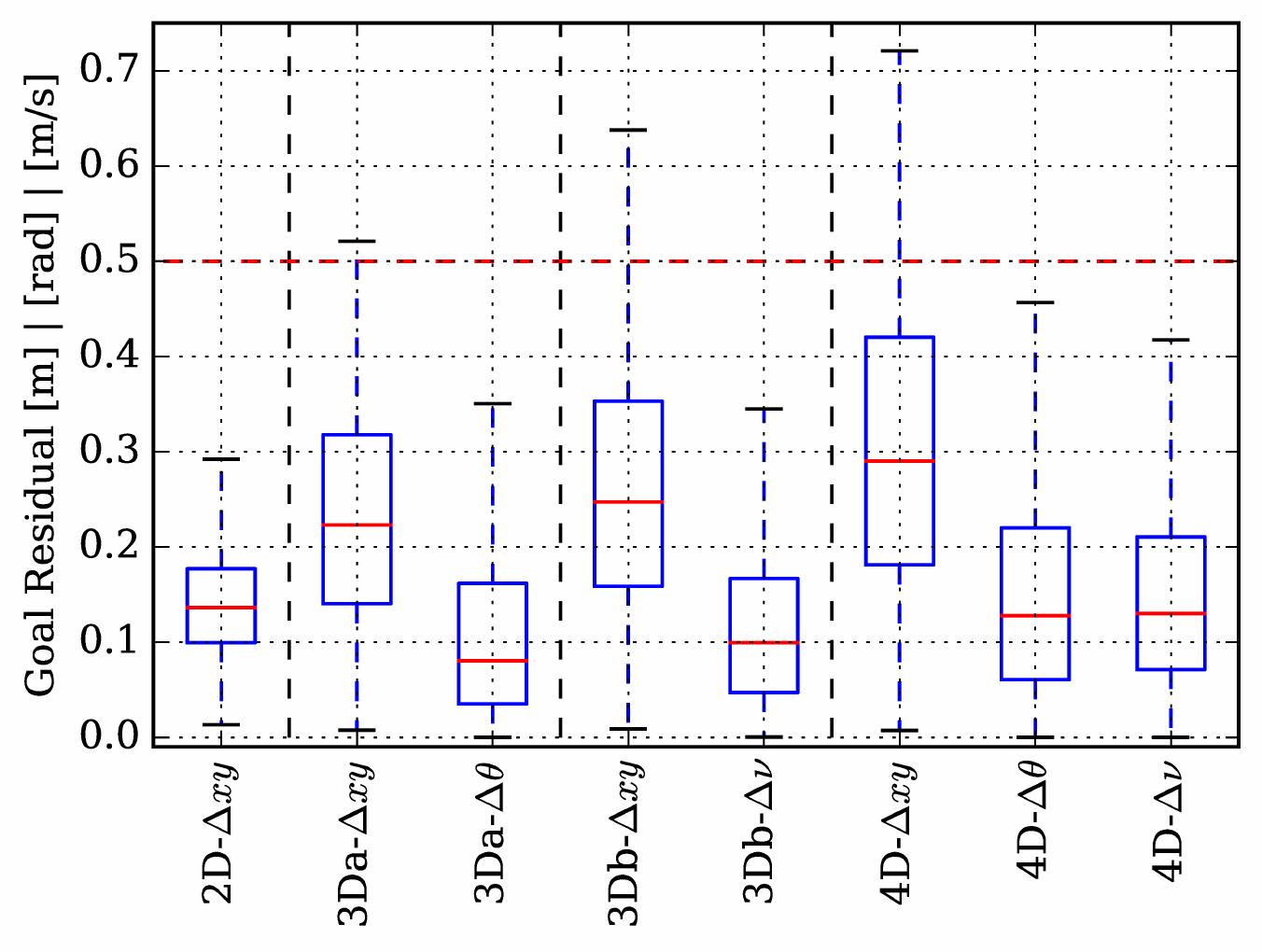}%
	\vspace{-4mm}%
	\caption{Accuracy per problem dimension.}%
	\label{fig:accuracy}%
\end{figure}%

We later evaluated the performance of the agents with randomly sampled start and goal state-pairs,
see Sect.~\ref{section:evaluation:simulation}. Fig.~\ref{fig:accuracy} shows the results for the four
agents. For each agent we separately plot the errors in the different dimensions.

The positional error is best for the 2D-problem (mean of $0.14$m) and increases as the problem
becomes more complex, i.e., a mean error of $0.23$m (3Da), $0.31$m (3Db), and $0.39$m (4D). The
mean angular error of $6.3\degree$ in the 3Da-problem also increases a bit to $10.3\degree$ for
the 4D-problem. The velocity error behaves similarly: from $0.14$m/s in 3Db is increases to
$0.20$m/s in the 4D-case. Please note, that the plot shows the median error, which is a bit
smaller.

What we do not see here is that the agents achieved success rates (i.e., that the error is
below a threshold of 0.5) of $100.0\%$ (2D), $99.9\%$ (3Da), $99.0\%$ (3Db) and even $97.6\%$
(4D) for a total number of 1,000 testing episodes. With increasing dimensions the problem
becomes more complex and harder to solve. However, the 4D-agent even solves the complex
motion planning task in most of the cases and satisfies all kinodynamic side-constraints at
any time


We also compared the time-efficiency of the generated trajectories to a cubic spline derived
from a velocity ramp. The spline's velocity starts from the initial velocity in the start state,
increases to the maximum possible velocity, and decreases to the target velocity at the latest
point. This achieves maximal linear acceleration and maximal velocity by design. But as the
spline interpolation does not consider kinematic constraints most of the generated trajectories
are not feasible. However, we use them to estimate a base-line for the duration of the generated
trajectories.

The mean and standard deviation of the duration ratio between the spline base-line and our agent
for the different subproblems was $(\mu=0.66;\sigma=0.04)$ (2D), $(0.81;0.08)$ (3Da),
$(0.81;0.26)$ (3Db), and $(1.29;0.55)$ (4D). Our RL-approach generates faster trajectories than
the spline-based approach in the 2D, 3Da and 3Db case. But the 4D-agent's trajectories take
$1.29$ times longer. However, this is the tradeoff between time-efficiency and kinodynamic
side-constraints as the trajectories of the RL-agent are feasible while the spline-based
trajectories are not.

\subsection{Application Results}
\label{section:evaluation:application}

\begin{figure}[t!]%
	\centering%
	\includegraphics[width=0.8\linewidth]{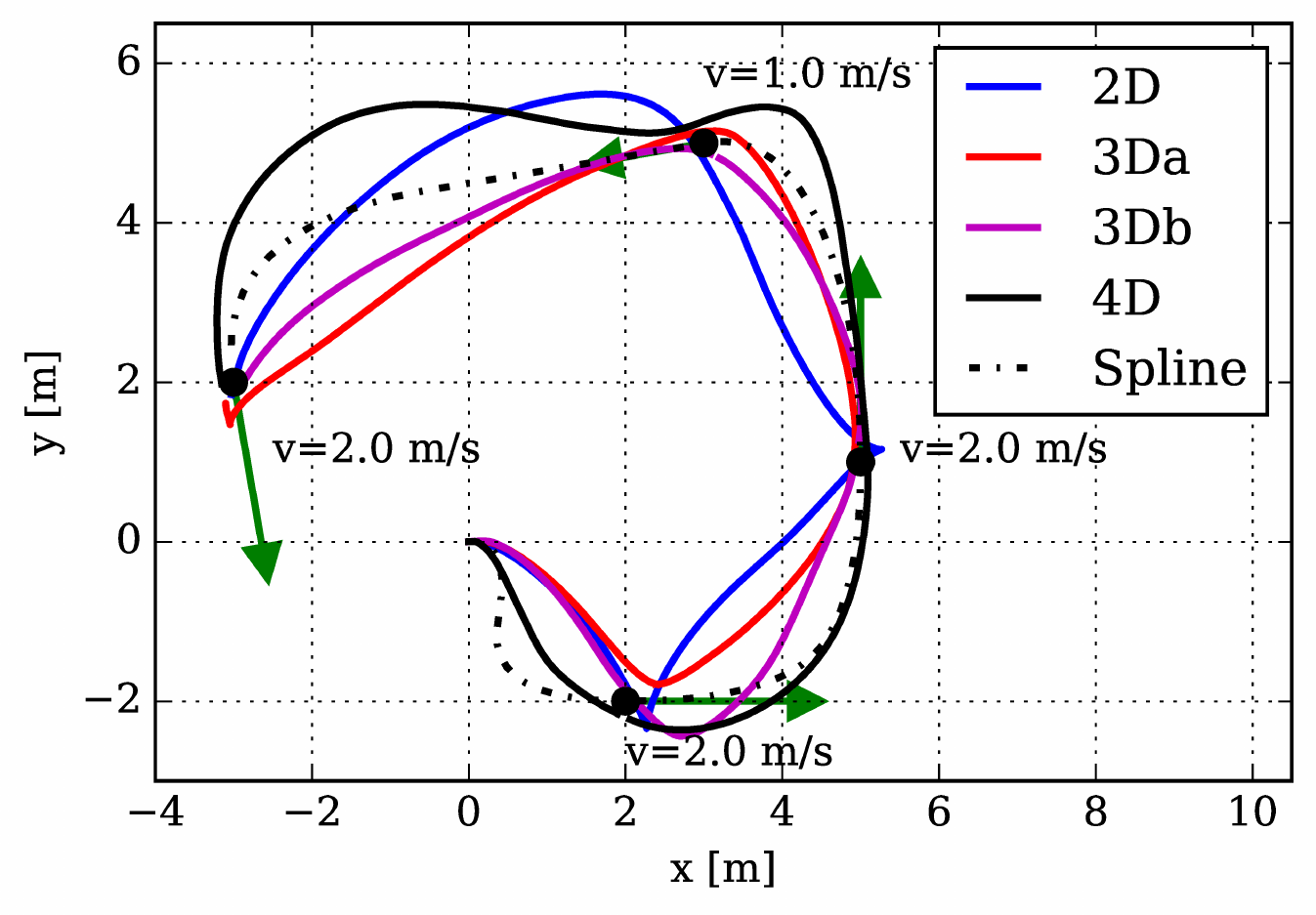}
	\vspace{-4mm}%
	\caption{Application test scenario.}%
	\vspace{-5mm}%
	\label{fig:eval:application}%
\end{figure}

To evaluate our approach for the use case we initially introduced we implemented a composite task
that continuously updates the goal-states, see Fig.~\ref{fig:eval:application}. We generate four
goal-states a-priori. The green vectors describe the states: the goal position is the start of the
vector, the length describes the velocity (also labeled) and the vector's orientation the goal
orientation of the robot at that state.
The agents did not see this scenario or such a situation within their training phase explicitly (although
similar combinations might have been included in the training episodes).

Fig.~\ref{fig:eval:application} shows the trajectories of the trained agents from
Sect.~\ref{section:evaluation:accuracy}. The dotted line denotes the spline-based
trajectory, which again, is not said to satisfy the constraints at any time. We see that the lower
dimensional agent D2 steers to the desired goal positions but generates
unsteady trajectories near the goal states. Considering the introductory use case, a motion
planner that generates trajectories only based on positions will fail to react fast enough (as
it introduces in-place rotations and zero velocities, see the first goal point). The 3Da-agent
behaves much more smoothly and generates correct orientations, but fails for the last goal point
where the orientation is achieved by an in-place rotation. The 3Db-agent is better and generates
a smooth trajectory. However, it may still also fail if the tactical analysis updates the goal
state that then might be at an unprofitable position.

But the 4D-agent not only gets close to the spline base-line but also considers all kinodynamic
constraints that are violated by the spline between the third and fourth goal-point.
The agent achieves a smooth and way-efficient steering to reach goal-points only by exploiting
the learned behavior.

\section{Conclusion}
\label{section:conclusion}

This papers presents a motion planning based on deep reinforcement learning. In contrast to previous
research in motion planning our algorithm does not compute trajectory plans a-priori or on-demand
but rather delivers a stream of control commands that we can use directly to steer a robot. Our
approach uses a reward function, state representation and a framework which we used to train an Deep
Deterministic Policy Gradient (DDPG) agent.

We evaluated our approach in a simulation environment and compared it
to a spline-interpolation. We also proved the applicability of our approach in a dynamic use-case that
continuously provides updated goal states.


\section*{Acknowledgements}
\label{section:acknowledgement}

This work was supported by the Bavarian Ministry for Economic Affairs, Infrastructure, Transport and Technology through the Center for Analytics-Data-Applications (ADA-Center) within the framework of "BAYERN DIGITAL II".

\bibliography{ref}
\bibliographystyle{abbrv}

\end{document}